%% file: 0-main.tex
\documentclass{article}

\usepackage[final]{corl_2022} 

\usepackage{babel}
\usepackage{collcell}
\usepackage{hhline}
\usepackage{pgf}
\usepackage{pdfx}
\usepackage{multirow}

\usepackage{wrapfig}
\usepackage{outlines}
\usepackage{microtype}
\usepackage{graphicx}

\usepackage{mathtools}
\usepackage{siunitx}
\usepackage[font=small,tableposition=top]{caption}
\usepackage{booktabs} 
\usepackage{amssymb}
\usepackage{amsmath,amsthm}
\usepackage{mathtools}
\usepackage{subcaption}
\usepackage{algorithm}
\usepackage[noend]{algpseudocode}
\usepackage[utf8]{inputenc}
\usepackage[inline]{enumitem}

\usepackage{dsfont}
\usepackage{hyperref}
\hypersetup{
    colorlinks=true,
    linkcolor=orange,
    filecolor=magenta,      
    urlcolor=orange,
    citecolor=orange,
}
\usepackage{cleveref}
\usepackage{changepage}

\usepackage{algorithmicx} 

\usepackage{titlesec}
\titlespacing\section{0pt}{0pt plus 2pt minus 2pt}{0pt plus 2pt minus 2pt}
\titlespacing\subsection{0pt}{3pt plus 4pt minus 2pt}{0pt plus 2pt minus 2pt}
\titlespacing\subsubsection{0pt}{3pt plus 4pt minus 2pt}{0pt plus 2pt minus 2pt}

\newcommand{\todo}[1]{}
\renewcommand{\todo}[1]{{\color{red} TODO: {#1}}}

\newcommand{\incmtt}[1]{{\fontfamily{cmtt}\selectfont{#1}}} 
\definecolor{darkblue}{rgb}{0.0,0.0,0.7} 

\algnewcommand{\LineComment}[1]{\textcolor{darkblue}{\scriptsize{\incmtt{\#\ #1}}}}

\title{Learning Visuo-Haptic Skewering Strategies \\ for Robot-Assisted Feeding}

\author{
 Priya Sundaresan\\
 Stanford University\\
 \texttt{priyasun@stanford.edu}
 \and
 \textbf{Suneel Belkhale}\\
 Stanford University\\
 \texttt{belkhale@stanford.edu}
 \and 
 \textbf{Dorsa Sadigh}\\
 Stanford University\\
 \texttt{dorsa@cs.stanford.edu}
 }

\makeatletter
\def\thanks#1{\protected@xdef\@thanks{\@thanks
        \protect\footnotetext{#1}}}
\makeatother

\begin{document}

\input{0-defs.tex}
\maketitle
\vspace{-0.7cm}
\begin{abstract}
Acquiring food items with a fork poses an immense challenge to a robot-assisted feeding system, due to the wide range of material properties and visual appearances present across food groups. Deformable foods necessitate different skewering strategies than firm ones, but inferring such characteristics for several previously unseen items on a plate remains nontrivial. Our key insight is to leverage visual and haptic observations during interaction with an item to rapidly and reactively plan skewering motions. We learn a generalizable, multimodal representation for a food item from raw sensory inputs which informs the optimal skewering strategy. Given this representation, we propose a zero-shot framework to sense visuo-haptic properties of a previously unseen item and reactively skewer it, all within a single interaction. Real-robot experiments with foods of varying levels of visual and textural diversity demonstrate that our multimodal policy outperforms baselines which do not exploit both visual and haptic cues or do not reactively plan. Across 6 plates of different food items, our proposed framework achieves 71\% success over 69 skewering attempts total. Supplementary material, datasets, code, and videos can be found on our \href{https://sites.google.com/view/hapticvisualnet-corl22/home}{website}.
\end{abstract}


\input{1-introduction.tex}

\input{2-related-work.tex}
\input{4-problem-statement.tex}

\input{CoRL 2022/5-method_revised}

\input{6-experiments.tex}
\input{7-conclusion.tex}

\clearpage

\footnotesize
\acknowledgments{
This work is supported by NSF Awards 2132847, 2006388, and 1941722 and the Office of Naval Research (ONR). Any opinions, findings, and conclusions or recommendations expressed in this material are those of the author(s) and do not necessarily reflect the views of the sponsors. Priya Sundaresan is supported by an NSF GRFP. We thank our colleagues for the helpful discussions and feedback, especially Tapomayukh Bhattacharjee, Jennifer Grannen, Lorenzo Shaikewitz, and Yilin Wu. }

\begin{small}
\bibliography{corl}  
\end{small}
\normalsize
\input{8-appendix.tex}

\end{document}

%% file: 0-defs.tex
\newcommand\smallO{
  \mathchoice
    {{\scriptstyle\mathcal{O}}}
    {{\scriptstyle\mathcal{O}}}
    {{\scriptscriptstyle\mathcal{O}}}
    {\scalebox{.6}{$\scriptscriptstyle\mathcal{O}$}}
  }

\def\colorModel{hsb} 

\newcommand\ColCell[1]{
  \pgfmathparse{#1<50?1:0}  
    \ifnum\pgfmathresult=0\relax\color{white}\fi
  \pgfmathsetmacro\compA{0}      
  \pgfmathsetmacro\compB{#1/100} 
  \pgfmathsetmacro\compC{1}      
  \edef\x{\noexpand\centering\noexpand\cellcolor[\colorModel]{\compA,\compB,\compC}}\x #1
  } 
\newcolumntype{E}{>{\collectcell\ColCell}m{0.4cm}<{\endcollectcell}}  
\newcommand*\rot{\rotatebox{90}}

%% file: 1-introduction.tex
\section{Introduction}

\label{sec:intro}
Realizing the full capabilities of assistive robots in the home, hospitals, or elderly care facilities remains challenging due to the dexterity required to complete many day-to-day tasks. Eating free-form meals is one such example with many nuances in perception and manipulation that can be easy to overlook. However, automating the task of feeding, one of six essential activites of daily life (ADL) \cite{katz1983assessing}, has the potential to improve quality of life for over one million people in the U.S. who are unable to feed themselves due to upper-extremity mobility impairment, their caregivers, families with young children and elders, and anyone impacted by the substantial time and effort required in meal preparation and feeding~\cite{brault2012americans, brose2010role, maheu2011evaluation}.

In recent years, there have been significant efforts to tackle the challenging problem of robot-assisted feeding. Solutions on the market have limited traction as they rely heavily on pre-programmed trajectories, pre-specified foods, have limited autonomy, or require manual utensil interchange~\cite{obi, mealmate}. Meanwhile, academic research on assistive feeding largely centers around data-driven methods but has yet to show widespread generalization across food groups~\cite{gallenberger2019transfer, feng2019robot, gordon2020adaptive, gordon2021leveraging}. As a necessary first step towards feeding, we focus on the problem of \emph{bite acquisition} --- acquiring bite-sized items from a plate or bowl --- using a robot with a fork-equipped end-effector. 
Developing a bite acquisition strategy sensitive to differences in \emph{geometry} and \emph{deformation} both \emph{across and within} food classes is a challenging problem: 
Skewering position and orientation matters for items with irregular shape, such as a broccoli floret where skewering at the stem is preferable to the head for stability of the acquisition. The fragility of food also affects the optimal skewering strategy, as delicate items such as thin banana slices are more likely to slip off a fork oriented vertically and instead benefit from an angled fork insertion and scooping strategy~\cite{bhattacharjee2019role}. On the other hand, hard baby carrots require a vertical insertion angle for effective and stable acquisition~\cite{bhattacharjee2019role}. In addition, instances within the same class of foods can also exhibit visual similarity but textural contrast (raw vs. boiled carrots, silken vs. extra firm tofu, cheddar vs. mozzarella cheese); choosing the optimal skewering strategy therefore depends on more than just vision.

Prior works show that food classification objectives can lead to visual features that can be used for downstream policy learning~\cite{gallenberger2019transfer, feng2019robot, gordon2020adaptive}. They also introduce an action taxonomy for skewering to discretize the complex space of possible acquisition trajectories~\cite{gallenberger2019transfer, feng2019robot, gordon2020adaptive, bhattacharjee2019towards, bhattacharjee2020more}.
Although these visual-only skewering strategies are able to classify food with different \emph{geometry}, they lack critical information about \emph{deformation} and may fail to differentiate between foods \emph{within} the same food class that appear similar but have drastically different properties, such as boiled and raw carrots. 

Preliminary experiments suggest that a boiled carrot requires the fork to skewer the item at an angle to avoid breakage or dropping during acquisition, whereas a rigid carrot requires a more forceful vertical approach to pierce the item. \citet{gordon2021leveraging} try to address this issue by leveraging post-hoc haptic feedback to update a visual policy after skewering. This work requires multiple trials of interaction per unseen food item to reason about item \emph{deformation} through haptic feedback. However, repeated skewering attempts can easily damage fragile items (e.g. overcooked carrots or thin slices of banana) and potentially change the properties of the food, leading to breakage or squishing over multiple robot interactions. These repeated interaction strategies do not scale to unseen food \textit{classes} either, for the same reasons. On the other hand, open-loop strategies that do not adapt skewering plans mid-motion are also limited in their ability to handle unseen items with unknown properties.

Therefore, bite acquisition methods should be able to \textit{zero-shot} generalize to new foods both within and across food classes, just like how humans skewer bites of food without the need for multiple interactions with the food.

Our key insight is to jointly fuse haptic and visual information during a single skewering interaction to learn a more robust and generalizable food item representation. We develop a bite acquisition system along with a visuo-haptic skewering policy that leverages this learned representation. 
The proposed representation informs the skewering policy of both \emph{geometry} and \emph{deformation} food properties on-the-fly, thus enabling a reactive policy which zero-shot generalizes to unseen plates of food \emph{within and across} food classes. Our experiments with a wide range of food items with varying geometry and deformability demonstrate that our method outperforms those that (1) do not jointly use haptic and visual cues and (2) do not reactively plan upon contact, achieving $71\%$ skewering success across $21$ items total. Our contributions include:
\vspace{-0.1cm}
\begin{itemize}[noitemsep,leftmargin=*]
    \item A skewering system that employs coarse-to-fine visual servoing to approach a food item, sense multimodal properties upon contact, and reactively plan skewering in the same continous interaction
    \item A zero-shot skewering policy that captures geometry and deformation by fusing visual and haptic information with demonstrated generalization to unseen food items
    \item Experimental validation on diverse seen and unseen food items, with varying degrees of visual likeness and deformation 
    \item An open-source dataset for multimodal food perception and custom end-effector mount designs, which we hope expands the scope of assistive feeding research
\end{itemize}

\begin{figure}[!tbp]
    \centering
    \includegraphics[width=1.0\textwidth]{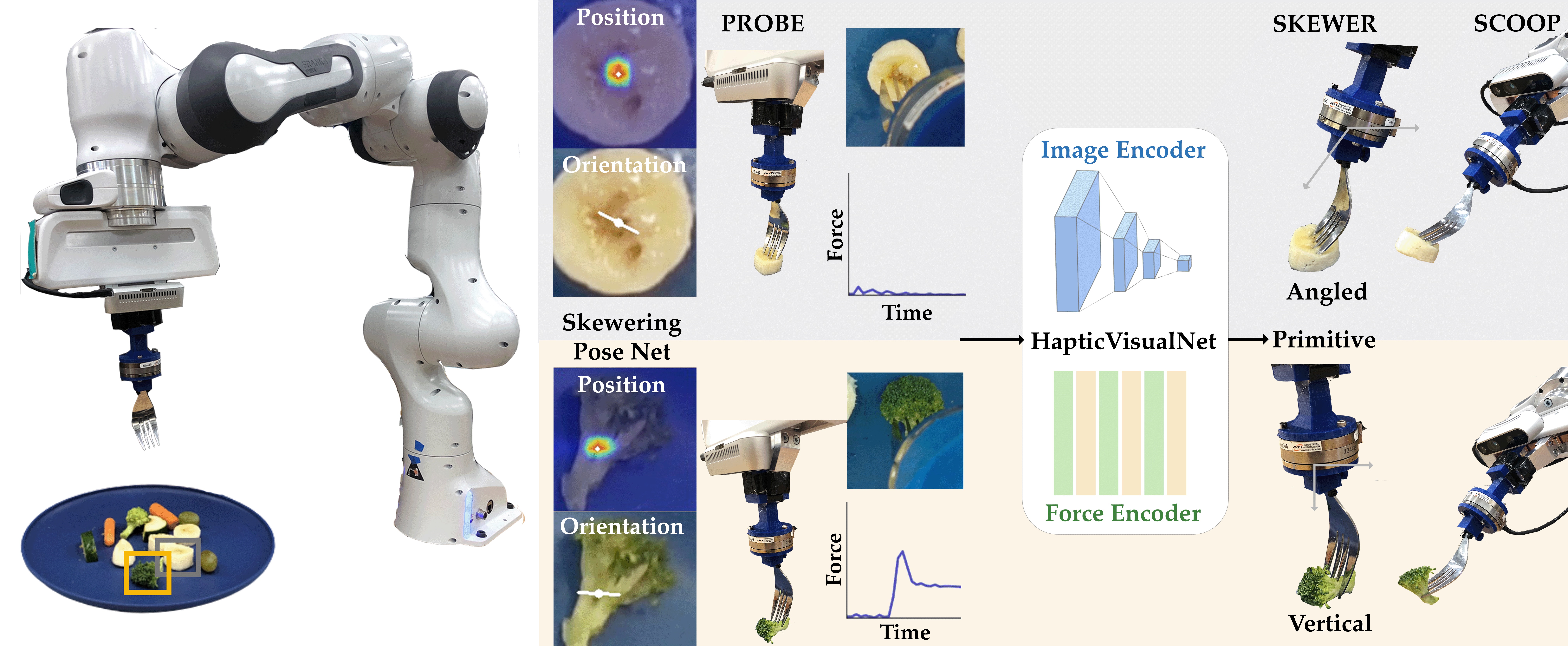}
    \caption{\emph{Left:} Our method learns zero-shot skewering of food items with a Franka Panda robot. Given an overhead plate observation, we localize food items and \emph{probe} them to reveal haptic and visual data. Using the multimodal data as input, HapticVisualNet infers the optimal skewering trajectory on the fly -- \emph{angled skewering} for soft items like banana slices, or  \emph{vertical skewering} for firm textures such as a broccoli stem.}
    \label{fig:splash_fig}
    \vspace{-0.9cm}
\end{figure}

%% file: 2-related-work.tex
\section{Related Work}
\label{sec:rw}
We build on prior works studying multisensory robot learning both within and beyond the food domain. In this section, we will discuss related work in robot-assisted feeding, food manipulation, and more generally multimodal robotic perception and manipulation.
\paragraph{Robot-Assisted Feeding}
Feeding can be split into two stages: bite acquisition and bite transfer. Previous work in bite transfer --- transferring an acquired bite to a user's mouth for consumption --- suggests that transfer is largely contingent upon acquisition~\cite{belkhale2022balancing, gallenberger2019transfer, bhattacharjee2020more}.
To enable reliable bite acquisition and thus transfer, recent acquisition frameworks combine image-based perception~\cite{gallenberger2019transfer, feng2019robot, gordon2020adaptive, gordon2021leveraging, bhattacharjee2020more}  --- bounding boxes and food pose estimates --- with an action space consisting of parameterized primitives that modulate fork roll/pitch relative to item geometry. SPANet (Skewering Position Action Network)~\cite{feng2019robot} is one such forward model mapping food image observations to actions, which has been shown to reasonably clear plates containing 16 types of fruits and vegetables. SPANet is trained on 2.5K fork interactions (81 hours of supervision~\cite{gordon2020adaptive, feng2019robot}) which does not readily scale to new foods. Follow-up works aim to rapidly adapt SPANet to unseen food items using a contextual bandit to learn the optimal primitive selection strategy from real interactions. Approaches include updating SPANet predictions online by observing the binary outcome of acquisition attempts on unseen items~\cite{gordon2020adaptive}, and additionally haptic time-series readings recorded post-hoc~\cite{gordon2021leveraging}. A key assumption in~\cite{gordon2021leveraging} is that the visual context observed \emph{pre-skewer} and the haptic context observed \emph{post-hoc} are equivalent alternate representations for the underlying food state, which does not always hold (e.g., firm and soft tofu appear to be almost identical but yield different haptic readings). In our work, we do not restrict ourselves to this assumption and consider the more general setting where visually similar items may have different physical properties. In addition, we do not assume access to repeated interaction trials with the food, and consider a zero-shot planning setting. ~\citet{bhattacharjee2019towards, song2019sensing} explore classifying food compliance or skewering outcomes from haptic data from a single interaction, but delegate reactive planning given these representations to future work. To address these gaps, we learn a multisensory policy that learns zero-shot skewering from \emph{pre-skewer} and \emph{post-contact} paired images and haptic readings.

\paragraph{Food Manipulation}
Recent simulated benchmarks for household food manipulation explore food preparation, lunch packing, and food storage~\cite{srivastava2022behavior}; pouring water~\cite{lin2020softgym}; pile manipulation for chopped food~\cite{suh2020surprising}; and drinking/feeding~\cite{erickson2020assistive}. While these works largely abstract away the state space of food, recent work in real robotic food slicing explores multimodal food representation learning from interaction using datasets consisting of paired visual, tactile, and audio information~\cite{gemici2014learning,zhang2019leveraging, sawhney2020playing, sharma2019learning, heiden2022disect}. In particular, \citet{gemici2014learning} propose to infer haptic properties (`brittleness', `tensile strength', `plasticity', etc.) from probing actions that can inform slicing actions, but does not consider visual properties. Likewise, \citet{zhang2019cutting} use vibrations from probing interactions to adapt slicing motions. The learned haptic representations from these work are not directly transferable to acquisition due to hardware and task differences, but we adopt the notion of using probing actions to inform skewering in our work.

\paragraph{Integrating Vision and Haptics in Robotics}
Vision-only manipulation has proven effective in robotic domains such as semantic grasping~\cite{mahler2017dex, florence2018dense} and deformable manipulation~\cite{ha2022flingbot}, but contact-rich tasks such as peg insertion~\cite{lee2019making} or robotic Jenga~\cite{fazeli2019see} have been shown to benefit from combining vision, force, and proprioception as inputs to a learned policy~\cite{li2019connecting, lee2021detect}. 
In light of these works, we propose to learn reactive, visuo-haptic policies for bite acquisition which remains largely unexplored. 
 

%% file: 4-problem-statement.tex


%% file: CoRL 2022/5-method_revised.tex
\section{Method}

\label{sec:method}
Our goal is to learn a multisensory manipulation policy which outputs skewering actions to clear a plate of bite-sized food without any previous skewering attempts for this food item. We consider foods of varying degrees of geometric, visual, and textural similarity, all on the same plate. We first formalize the bite acquisition setting in Section \ref{subsec:ps} and introduce our action space to tackle the problem in Section \ref{subsec:actions}. Next, we discuss our interaction protocol for sensing visuo-haptic properties of food (Section \ref{subsec:probe}), enabling us to learn a multimodal skewering policy (Sections \ref{subsec:multimod_repr}-\ref{subsec:training_data}).

\subsection{Problem Formulation}
\label{subsec:ps}

\begin{wrapfigure}{r}{0.3\textwidth}
  \vspace{-0.5cm}
  \centering
    \includegraphics[width=0.3\textwidth, trim=10pt 10pt 10pt 0]{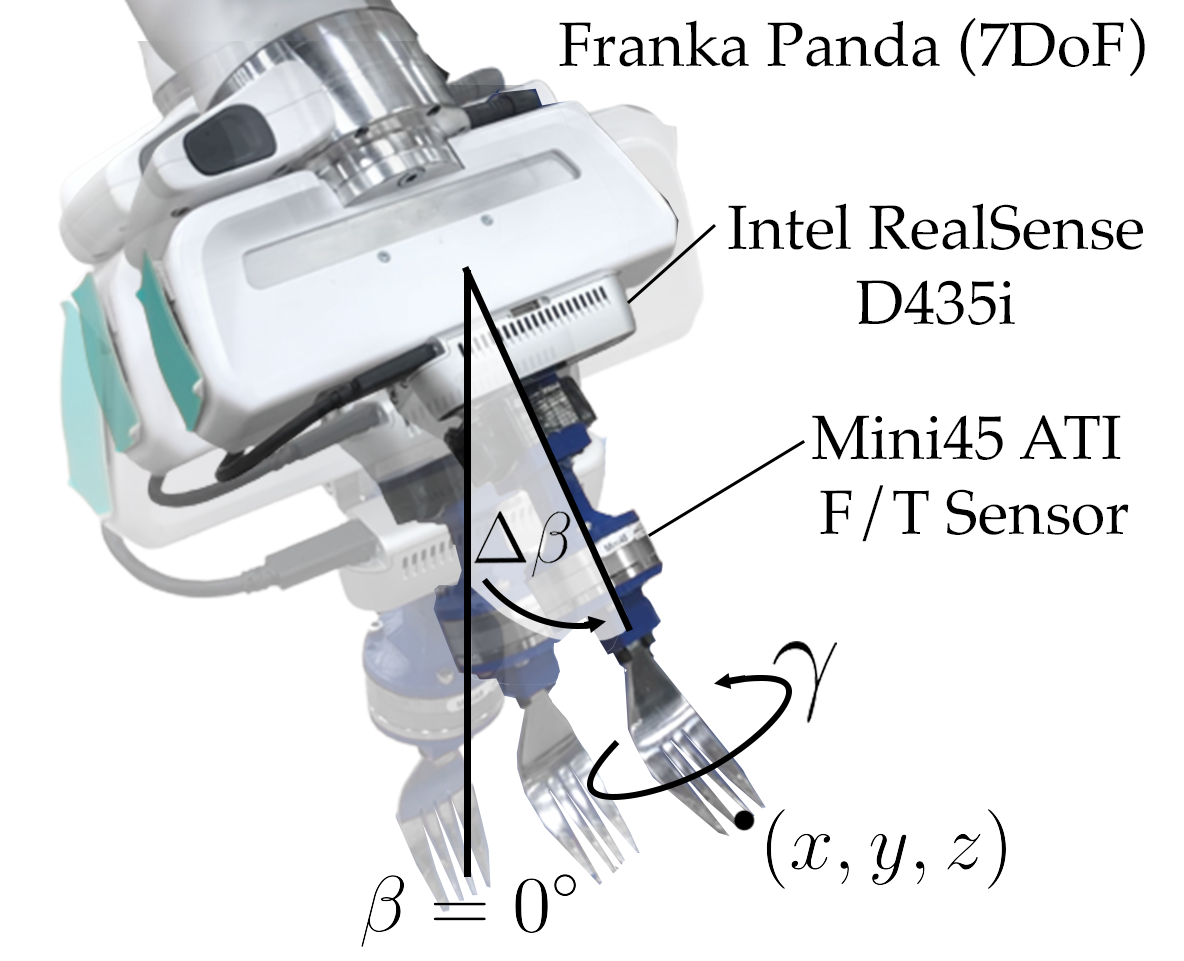}
  \caption{Skewering action space: fork pitch $\beta$, roll $\gamma$, and position $(x,y,z).$}
  \label{fig:action_space}
  \vspace{-0.1cm}
\end{wrapfigure}
At each timestep $t \in 1, \hdots, T$, we assume access to a the current RGB-D image observation $I_t \in \mathbb{R}^{W \times H \times C}$ of a plate of food and an $N$-length history of haptic readings $H_t \in \mathbb{R}^{N \times 6}$, denoting 6-axis readings from an F/T sensor on a fork-mounted end-effector. Similar to prior work, we consider an action space parameterized by fork $(x,y,z)$ position, roll $\gamma$, and pitch $\beta$. We define an action $a_t \in \mathcal{A}$, visualized in Fig.~\ref{fig:splash_fig}, as follows~\cite{gordon2020adaptive, gordon2021leveraging}: 
\begin{align}
\label{eq:action}
    a_t = (x,y,z,\Delta z,\gamma,\Delta \beta)
\end{align}
At time $t$, the fork starts in position $(x,y,z)$ with pitch $\beta=0^{\circ}$ and roll $\gamma$ and moves downward $\Delta z < 0$ while optionally tilting $\Delta \beta \geq 0^{\circ}$ to skewer an item (Figure \ref{fig:action_space}). 

We use $l_t[a_t] \in \{0,1\}$ to denote the binary loss of executing action $a_t$, where $l_t[a_t] = 1$ denotes failure, for example food failing to be picked up, slipping off the fork after skewering, or breakage or damage as a result of skewering. We opt for a binary loss objective as it is difficult to quantify more nuanced notions of success such as degree of damage to an item or stability of a skewer.


We aim to learn a policy $\pi_\theta(a_t\mid I_t, H_t)$ that minimizes $\sum_{t}^Tl_t[a_t]$ given no previous interactions. Our policy learns to map visual and haptic information to a discrete set of skewering primitives using a small but diverse labelled dataset that can be collected offline. By conditioning the policy on haptic readings during a short initial contact period with the the food item, our method HapticVisualNet can extract food properties that are inaccessible to vision using only a single skewering interaction. First we outline how we represent the discrete set of primitives for skewering.


\subsection{Skewering Action Primitive Parameterization}
\label{subsec:actions}

To successfully skewer a food item, the fork position and roll must align with the location and orientation of a food item. The fork pitch must also adapt to the compliance of a food item (i.e. a soft banana slice favors an \emph{angled} fork approach to prevent slip, while a raw carrot favors a \emph{vertical} approach for piercing). Thus, our action space employs two primitives, \texttt{vertical skewer} or \texttt{angled skewer}, to account for rigid or compliant items, respectively. We implement \texttt{vertical skewer} with $\Delta \beta = 0^{\circ}$, denoting no tilt during skewering, and \texttt{angled skewer} with $\Delta \beta > 0^{\circ}$, where the fork gradually tilts from vertical to an angled approach during insertion into a fragile food item (Figure \ref{fig:action_space}). Prior work includes both angled and vertical skewering strategies amongst an even larger set of primitives, but we empirically find that our simplified taxonomy reduces redundancy in this larger action space and can handle an equivalently broad range of food items, evaluated in Section \ref{sec:evaluation} ~\cite{feng2019robot, gordon2021leveraging}. 
In order to decide between these strategies on-the-fly, we condition our policy not only on visual information but also on haptic information at the point of contact with the food item, discussed in the next section.
 
\subsection{Sensing Multimodal Data Via Probing}
\label{subsec:probe}
We introduce a \emph{probing} motion to obtain visuo-haptic information about a food item by bringing the fork in contact with the item surface but without actually skewering it. The multisensory information collected from probing serves as input to $\pi_\theta$ which rapidly decides the skewering primitive to execute. Between the probing and skewering phases, the fork remains stationary and in constant contact with the food item, thus enabling a fluid transition between phases.

Our \emph{probe-then-skewer} approach requires localizing an item and approaching with precision so as to not accidentally shift or topple it while making contact. To accomplish this, we first detect items from a plate image using a pre-trained RetinaNet food bounding box detector from~\citet{gallenberger2019transfer}. Similar to prior work, we also train a network (SkeweringPoseNet) which refines the estimated item location by predicting a keypoint for the item center within the local bounding box, and a fork roll angle $\hat{\gamma}$ with which to approach~\cite{feng2019robot}. We can obtain the 3D predicted item location $(\hat{x}, \hat{y}, \hat{z})$ in robot frame by using depth information. Next, we continuously servo to the item using a learned model (ServoNet) which detects the fork-item offset as keypoints from streamed RGB images. Using this framework, we probe a food item and record a post-contact image observation $I_t$ and the short window of force magnitude readings $H_t$. In Section \ref{subsec:multimod_repr}, we discuss how these multisensory readings inform the optimal skewering strategy.

\subsection{Multimodal Representation and Policy Learning}
\label{subsec:multimod_repr}
To learn $\pi_\theta(a_t\mid I_t, H_t)$, we propose HapticVisualNet, a network which takes observations captured by the probing motion and reactively outputs the appropriate skewering action. HapticVisualNet takes as input a post-contact image of a food item $I_t \in \mathbb{R}^{W \times H \times 3}$ and $H_t$, the force magnitude readings recorded from the F/T sensor during the first $N$ milliseconds of contact in the initial probing period.

HapticVisualNet maps $(I_t, H_t)$ to an $|\mathcal{A}|$-d vector denoting the likelihood of success for each action primitive, in our case \texttt{vertical skewer} and \texttt{angled skewer}. The model first encodes visual information and haptic information separately, and then concatenates these features to produce a joint visuo-haptic representation. The policy then predicts action success likelihood from this representation. We implement HapticVisualNet as a multi-headed network with a ResNet-18 backbone for the visual encoder and a LSTM for the haptic encoder. We pass the concatenated visual and haptic encodings to a linear layer followed by a softmax to obtain primitive successes, and choose the maximum likelihood predicted primitive as the skewering action.

\subsection{Training and Data Collection}
\label{subsec:training_data}

We train HapticVisualNet on a small but diverse dataset of 300 paired post-contact images and haptic readings, augmented 8x using image affine and colorspace transforms as well as temporal scaling and shifting of the haptic readings. The dataset consists of \emph{hard} items labeled \texttt{vertical skewer} (raw carrots / broccoli / zucchini / butternut squash, grapes, cheddar cheese, and celery) and \emph{soft} items labeled \texttt{angled skewer} (banana, kiwi, ripe mango, boiled carrots / broccoli / zucchini / butternut squash, avocado, and mozzarella cheese).

In practice, we use an $N=$ 26 ms. contact window of haptic readings, from the initial probing period, which we find adequately captures force-surface interactions. Using a 20-30 ms. window of contact is also a common choice of haptic representation in other reactive, contact-rich manipulation settings~\cite{lee2019making, gordon2021leveraging}. For each paired example, we manually assign the optimal primitive label --- \texttt{vertical} or \texttt{angled skewer} --- based on whether the annotator considers the item hard or soft. This process requires ~3 hours of data collection and labeling time total (a 27x reduction from the 81 hours reported in SPANet~\cite{gordon2021leveraging}), and \emph{without} the need for actual skewering attempts during data collection. We intend for the inclusion of haptic data to prevent overfitting to visual features, enabling a more generalizable, food representation trained with greater sample efficiency. Both ServoNet and SkeweringPoseNet are implemented using a ResNet-18 backbone, each trained on 200 images of the same items (2 hours of supervision) and augmented to a dataset of 3,500 paired images and annotations. Additional training and implementation details are in Appendix \ref{sec:servonet_details}.

In summary, our method leverages offline datasets to learn visuo-haptic features to more robustly predict between a set of discrete skewering actions. The probing motion used to obtain haptic information is connected seamlessly to the chosen skewering motion, leading to one continuous and adaptive zero-shot skewering policy aware of haptics \textit{and} vision.

%% file: 6-experiments.tex
\section{Evaluation}
\label{sec:evaluation}

\begin{figure}[!htbp]
    \centering
    \includegraphics[width=1.0\textwidth]{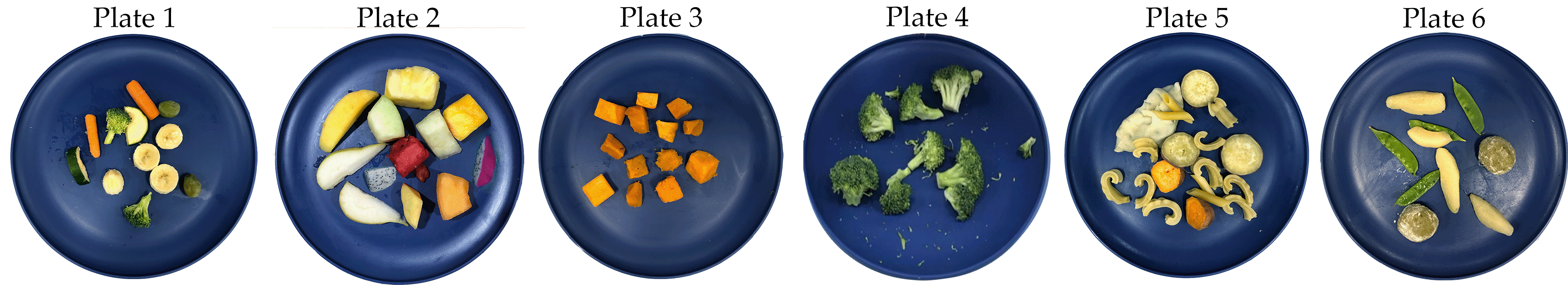}
    \caption{6 Plates for Evaluation, covering a wide range of foods. From left to right: \textbf{1}: Raw banana, broccoli, zucchini, carrot, grapes, cucumber. \textbf{2 (Unseen)}: Raw pineapple, mango, dragonfruit, canteloupe, honeydew, pear. \textbf{3}: Raw butternut squash, boiled butternut squash. \textbf{4}: Raw broccoli florets. \textbf{5 (Unseen)}: Pasta, dumpling, boiled yam/sweet potato, raw yam/sweet potato. \textbf{6 (Unseen)}: Ice cream mochi, snow peas, canned peaches.}
    \label{fig:exps}
    \vspace{-0.3cm}
\end{figure}

\begin{figure}[!htbp]
    \centering
    \includegraphics[width=1.0\textwidth]{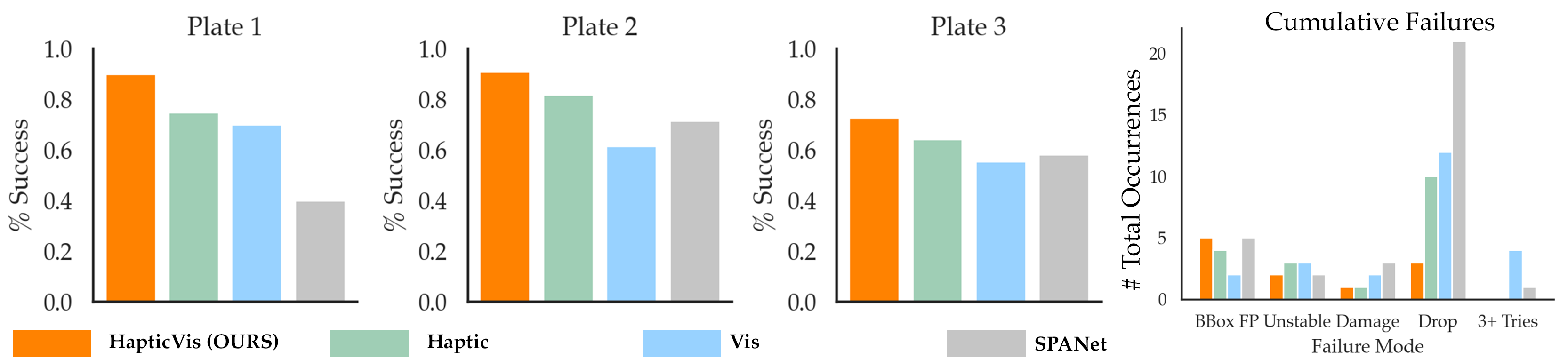}
    \caption{\textbf{Skewering Success and Failure Modes}: We visualize the number of items acquired over total acquisition attempts for all methods across Plates 1-3. Failure modes include exceeding the maximum number of consecutive attempts (3+ tries per item), dropping after skewering, the item being unstable on the fork after skewering (affecting transfer), damage or breakage to fragile items, or failure to detect an item due to bounding box anomalies. HapticVisNet (ours) performs best on each plate, while causing the least failures.}
    \label{fig:exps}
    \vspace{-0.5cm}
\end{figure}

In this section, we seek to evaluate (1) the benefits of combining both vision and haptics for bite acquisition as opposed to only using a subset of modalities, (2) the effectiveness of reactive skewering compared to open-loop strategies, and (3) the generalization capabilities of our system to previously unseen foods. We first perform classification ablations of HapticVisualNet in (Section \ref{subsec:ablations}) and then deploy our system in the real world for trials on both seen and unseen foods (Section \ref{subsec:hardware}-\ref{subsec:real_exps}).

\subsection{Ablative Studies}
\label{subsec:ablations}

We evaluate the contributions of both visual and haptic data towards primitive classification by training and evaluating HapticVisualNet against two variants which observe exclusively the post-contact haptic readings or post-contact image after probing. 
 
 

\begin{wrapfigure}{r}{0.52\textwidth}
  \vspace{-0.25cm}
  \centering
    \includegraphics[width=0.52\textwidth,trim=40pt 0 40pt 40pt]{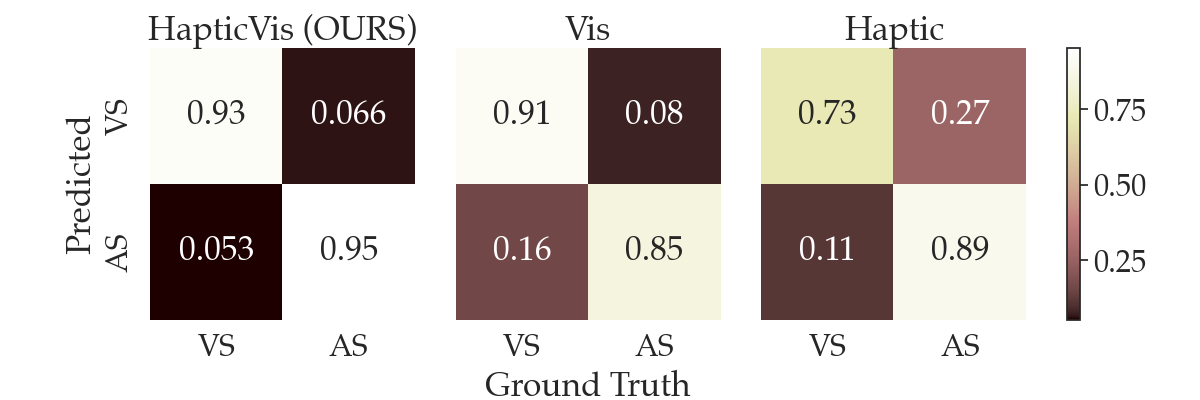}
  \caption{Confusion matrices for classification accuracy for the skewering primitive (\texttt{angled skewer}  or \texttt{vertical skewer}), on a held-out test dataset of 60 images, for each model. Darker off-diagonals and lighter on-diagonals indicate more accurate models.}
  \label{fig:confusion_mat}
  \vspace{-0.5cm}
\end{wrapfigure}
Figure \ref{fig:confusion_mat} shows the confusion matrices for classification accuracy. HapticVisualNet achieves the highest overall and per-class classification accuracy, and omitting either modality hurts accuracy. We hypothesize that the reduced performance of the vision-only model stems from visually similar but texturally dissimilar foods in our dataset, for which inferring the primitive is challenging without haptic context. On the other hand, the haptic-only model learns a na\"ive solution of mapping high magnitude contact events to \texttt{vertical skewering} and low readings to \texttt{angled skewer}. This is brittle for anomalous foods like broccoli which benefit from a \texttt{vertical skewer}, yet may yield low contact readings if the fork comes in contact at the head instead of the stem. Similarly, the fork can easily penetrate a thin banana slice and touch the plate during probing, yielding high contact readings when an \texttt{angled skewer} is still preferable. In the subsequent sections, we evaluate HapticVisualNet (HapticVis) against the haptic only (Haptic), vision only (Vis), and SPANet baselines on real food acquisition trials. See Appendix \ref{sec:ablations_rebuttal} for additional ablations of the learned multimodal representation and HapticVisualNet sample efficiency.

\subsection{Hardware Setup}
\label{subsec:hardware}
Our setup consists of a 7DoF Franka Emika Panda robot with the default gripper. We outfit the gripper with a custom 3D-printed mount comprised of a standard fork, a Mini45 ATI F/T sensor, and a D435i RealSense camera. We perform all acquisition trials on a plastic dinner plate on an anti-slip surface. We instantiate each primitive, parameterized according to Equation \eqref{eq:action} as follows, assuming a fixed $\Delta z$ and discretized $\Delta \beta$: 
\begin{itemize}
    \item \texttt{probe} = $(\hat{x}, \hat{y}, \hat{z}$-\text{\texttt{APPROACH\_HEIGHT}}, -\text{\texttt{APPROACH\_HEIGHT}}, $\hat{\gamma}, 0^{\circ})$
    \item \texttt{vertical skewer} = $(\hat{x}, \hat{y}, \hat{z}$, -\text{\texttt{DT*0.17m/sec}}, $\hat{\gamma}, 0^{\circ})$
     \item \texttt{angled skewer} = $(\hat{x}, \hat{y}, \hat{z}$, -\text{\texttt{DT*0.08m/sec}}, $\hat{\gamma}, 65^{\circ})$
\end{itemize}

Here, $(\hat{x}, \hat{y}, \hat{z})$ denotes a predicted food item location, obtained by deprojecting a predicted pixel in a depth image from ServoNet to a 3D location. SkeweringPoseNet also predicts the fork roll $\hat{\gamma}$. We first \texttt{probe} the item starting from an \texttt{APPROACH\_HEIGHT} of 0.01cm, observe a post-contact image $I_t$, and record haptic readings $H_t \in \mathbb{R}^{26}$ over the first 26-milliseconds of contact. Given these inputs, HapticVisualNet infers either \texttt{vertical skewer} or \texttt{angled skewer} which we execute.

The robot controller runs at 20Hz (\texttt{DT} = $0.05$), and primitives terminate early if the fork reaches a pre-defined $z$-distance (the plate height) or force limit. After skewering, the end-effector \emph{scoops} until the fork is nearly horizontal with $\beta = 80^{\circ}$, emulating the start of a feasible \emph{transfer} trajectory. 

\subsection{Baselines}
We deploy all methods --- HapticVis (ours), Haptic, Vis, and SPANet --- on the real robot setup of Section \ref{subsec:hardware} and classification networks trained according to Section \ref{subsec:ablations}. The HapticVis, Haptic, and Vis methods all perform \emph{probing-then-skewering}, but run inference using both, only haptic, or only visual sensory information obtained from probing, respectively. 

We also implement SPANet given the pre-trained visual models and original taxonomy of six skewering primitives reported in ~\cite{feng2019robot}. SPANet performs zero-shot primitive inference given an overhead image observation of a food item, without probing. SPANet still uses our ServoNet to plan, analogous to the original implementation which similarly accounted for fork-food precision error.

\subsection{Real World Bite Acquisition Results}
\label{subsec:real_exps}
We compare all methods on the challenging task of clearing plates containing 10 bite-sized food items, evaluated according to skewering success and the distribution of skewering failure modes encountered (Table \ref{table:phys_exp_results}). We define a skewering success as one in which the fork picks up the item with at least 2 tines inserted and the item remains on the fork for up to 5 seconds after scooping as in~\cite{feng2019robot}. Failure modes are detailed in Figure (\ref{fig:exps}). Between successful acquisitions, a human operator removes the acquired item from the fork. Upon skewering failure, an attempted item remains on the plate and can be re-attempted up to 3 times before being manually removed and marked as a failure.  

\smallskip \noindent \textbf{Plates 1-3 -- Full System Evaluation}:
We first perform a full-system evaluation of HapticVis and all baselines on 3 plates. Plate 1 contains in-distribution fruits and vegetables that HapticVisualNet was trained on which include both textural and visual diversity, Plate 2 consists of unseen assorted fruits with visual diversity but similar textures, and Plate 3 contains in-distribution boiled and raw butternut squash cubes which appear similar but greatly differ in softness. When adjusted for perception failures (e.g., bounding box false negatives) which affect all methods, HapticVis outperforms all methods across Plates 1-3 (Figure \ref{fig:exps}). The bulk of HapticVis failures center around near-misses or perception failures which are less drastic than damaging items or exceeding skewering attempts (Figure \ref{fig:exps}). Vis slightly underperforms Haptic across all plates and achieves lowest performance on Plate 3, where mispredicted \texttt{vertical skewer} or \texttt{angled skewer} actions can miss/damage soft-boiled squash, or fail to penetrate hard raw squash. SPANet achieves lowest performance, mostly due to erroneously executing vertical strategies to pick up soft items like bananas and ripe dragonfruit in many cases. SPANet's underperformance relative to HapticVis, Haptic, and Vis suggests the effectiveness of reactive strategies compared to open-loop acquisition, and that our simplified action space is just as expressive as SPANet's larger taxonomy. We acknowledge that the performance gap may in part be attributed to hardware differences (different robot, different fork mount) in the original SPANet compared to our re-implementation. 

\smallskip \noindent \textbf{Plate 4 -- Texturally Misleading Food}: While Haptic is the most competitive baseline on Plates 1-3, we run additional experiments between HapticVis and Haptic on a plate of only broccoli florets (Plate 4). In cases where the stem is occluded from view or the fork servos to the leafy region to make contact, Haptic tends towards misclassifying the low readings as \texttt{angled skewer}, leading to frequent failures to pierce the item which HapticVis is better equipped to recognize and avoid.

\smallskip \noindent \textbf{Plates 5-6 -- Generalization to Out-of-Distribution Foods}:
Finally, we stress-test the generalization capabilities of HapticVis on two plates of unseen foods, Plate 5 and Plate 6, containing the items listed in Table \ref{table:phys_exp_results}. Across Plates 5-6, HapticVis achieves 58\% success, a 19\% improvement over Vis, indicating the promise of multimodal representations for zero-shot food skewering. HapticVis performs best on soft canned pears and boiled/raw root vegetables which are most comparable to the items in the training distribution, closely followed by pasta and ice cream mochi. The majority of failures occur due to the fragility of dumplings and thinness of snow peas which are difficult to pierce. Still, by fusing haptic and visual information, HapticVis is better equipped to generalize to visually and texturally diverse foods. See Appendix \ref{sec:stress_tests} for additional stress-tests of HapticVisualNet.

\begin{table}[!htbp]
\centering
\renewcommand{\arraystretch}{1.7}
\setlength{\tabcolsep}{2pt}
\fontsize{10}{7}\selectfont
\vspace{-0.2cm}
\resizebox{\columnwidth}{!}{
 \begin{tabular}{c  c  c  c  c | c  c  c  c } 
 \multicolumn{5}{c}{\textbf{Plate Type}} &
 \multicolumn{4}{c}{\textbf{\# Items Acquired / Total Attempts}} \\ 
  \hline
\textbf{\emph{Plate}} & \textbf{\emph{Items}} & \textbf{\emph{Visuals}} & \textbf{\emph{Haptics}} & \textbf{\emph{Category}} & \textbf{\emph{HapticVis}} & \textbf{\emph{Haptic}} & 
\textbf{\emph{Vis}} & 
\textbf{\emph{SPANet}} \\ 
\hline
1 & Assorted fruits and vegetables & Diverse & Diverse & Seen & \textbf{9/10} & 9/12 & 7/10 & 8/20 \\
2 & Assorted tropical fruits & Diverse & Similar & Unseen & \textbf{10/11} & 9/11 & 8/13 & 10/14 \\
3 & Boiled/raw butternut squash & Similar & Diverse & Seen & \textbf{8/11} & 9/14 & 10/18 & 7/12 \\
4 & Broccoli florets & Similar & Similar & Seen & \textbf{8/13} & 7/17 & -- & -- \\
5 & Pasta, dumplings, boiled/raw root veggies & Diverse & Diverse & Unseen & 7/11 & -- & \textbf{9/13} & --\\
6 & Mochi, snow peas, canned pear & Diverse & Diverse & Unseen &  \textbf{7/13} & -- & 5/23 & --\\
\cline{6-9}

 \multicolumn{5}{c|}{} & \textbf{71\%} & 63\% & 51\% & 54\%  

 \end{tabular}}
 \vspace{-0.1cm}

\caption{\textbf{Physical Results:} We evaluate the ability of all methods to clear 6 plates (Figures \ref{fig:exps}), each initially containing 10 items. The number of items acquired refers to items successfully skewered ($\leq 10$ in all cases due to dropped items or early termination with bounding box false negatives). The total attempts refers to all attempted acquisitions until termination or clearance ($\geq 10$ in all cases due to failed items remaining on the plate for up to 3 consecutive re-attempts). HapticVis outperforms baselines in 5/6 plates. The symbol -- denotes experiments that are not useful in the specific testing scenario.}
\label{table:phys_exp_results}
\vspace{-0.3cm}
 \end{table}

Overall, HapticVisualNet benefits from the use of both vision and haptics from just a single interaction, and outperforms single-modality baselines for a variety of challenging food plates. We show that skewering strategies that reactively update the strategy (HapticVis, Haptic, Vis) are more robust to texturally and visually diverse food items than open loop strategies (SPANet). With its multimodal representation, HapticVisualNet can also zero-shot generalize to challenging unseen food classes.
 

%% file: 7-conclusion.tex
\section{Discussion}
\label{sec:conclusion}
\smallskip \textbf{Summary} In this work, we present a framework for zero-shot food acquisition of a diverse range of bite-sized items using multimodal representation learning. Our approach uses interactive probing to sense complex, multisensory food properties upon contact in order to reactively plan a skewering strategy. We deploy the learned policy along with a learned visual servoing controller on a robot for zero-shot skewering of unseen food items. Our experiments span a wide range of food appearances and textures, and validate the need for multimodal reasoning and reactive planning to clear plates.

\smallskip \textbf{Limitations and Future Work} One limitation in our approach is the use of a small action space for acquisition. While our set of primitives can generalize to a wide range of foods, our current strategies are not equipped to handle thin, flat items like finely sliced produce or leafy salad greens, which may require new techniques like positioning the fork under an item to scoop, gathering multiple items together before skewering, or using plate walls for stabilization. Other food groups like noodles would benefit twirling, and foods with irregular geometries might require toppling into a stable pose before skewering. Another limitation in this work is the supervision used to currently train our multimodal policy. However, since our policy learns from sparse labels, we are excited by the possibility of automatically detecting skewering outcomes to self-supervise the training procedure for HapticVisualNet. Finally, in future work we hope to tackle challenging food groups such as filled dumplings which easily break and extremely thin snow peas which are hard to pierce.

%% file: 8-appendix.tex
\newpage
\appendix
\begin{LARGE}
\begin{center}
\textbf{Learning Visuo-Haptic Skewering Strategies for Robot-Assisted Feeding \\ Supplementary Material}
\end{center}
\end{LARGE}

Datasets, code, and videos can be found on our \href{https://sites.google.com/view/hapticvisualnet-corl22/home}{website}.

\section{Experimental Hardware}
\begin{figure}[!htbp]
    \centering
    \includegraphics[width=1.0\textwidth]{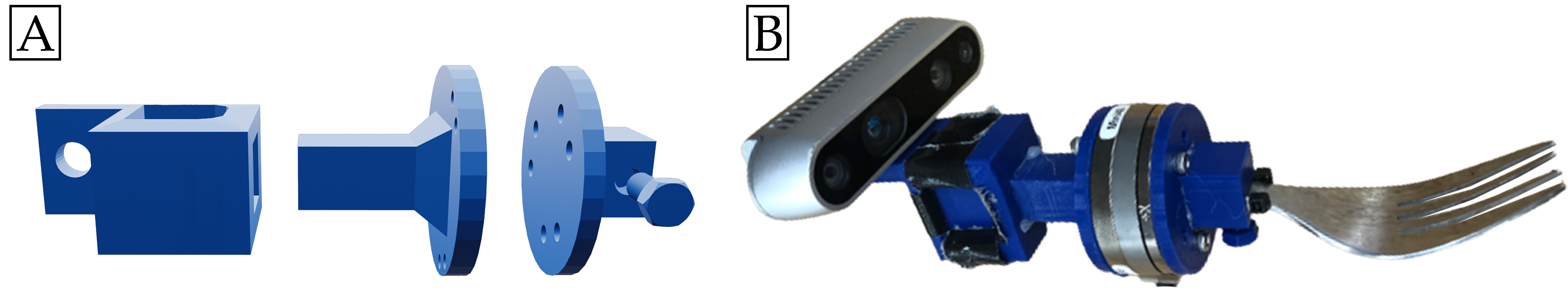}
    \caption{\textbf{End-Effector Custom Hardware:} A) CAD designs for custom end-effector attachment with inserts for Franka Panda gripper tips and mounting holes for Intel RealSense D435i, Mini45 ATI F/T sensor, and fork with screw attachment. B) The 3/D printed full mount, assembled using mounting screws and super glue.}
    \label{fig:end_effector_mount}
    \vspace{-0.2cm}
\end{figure}

\begin{figure}[!htbp]
    \centering
    \includegraphics[width=0.9\textwidth]{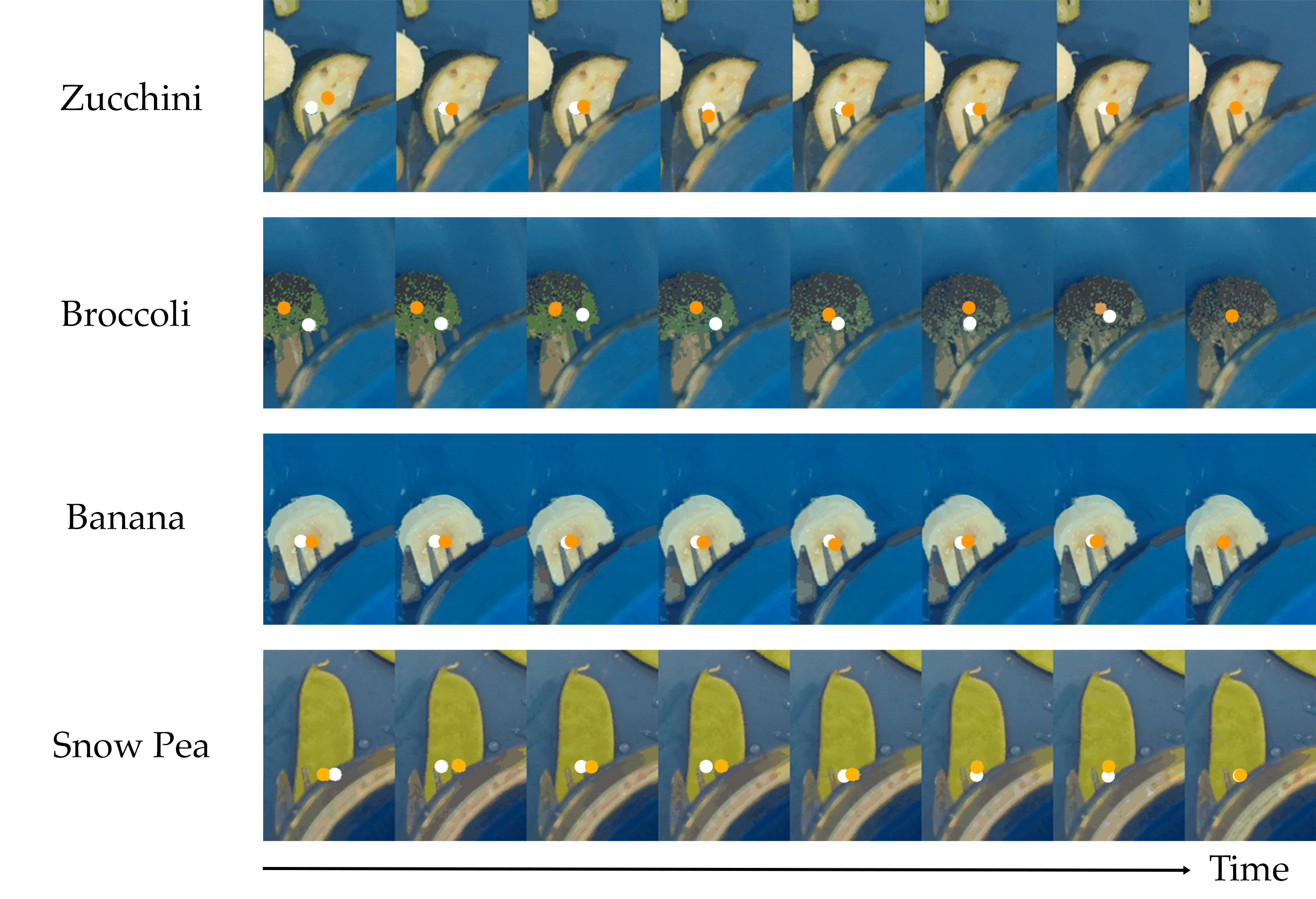}
    \caption{\textbf{ServoNet}: From left to right, we plot the last 8 frames of visual servoing using ServoNet until the fork tines midpoint (white) is centered with the predicted item location (orange), across 4 items (zucchini, broccoli, banana slice, snow pea).}
    \label{fig:servo_net}
\end{figure}

We collect all data and run all experiments using the Franka Panda 7DoF robot with the custom 3D-printed end-effector mount shown in Figure \ref{fig:end_effector_mount}. While the fork is attached via a screw-in slot, the force of consecutive skewering attempts in the course of clearing a plate can cause the fork position to shift slightly within the slot. To address this, we train ServoNet to estimate the fork-food item offset and continuously servo until the midpoint of the tines is aligned with the predicted item center (Figure \ref{fig:servo_net}).

\section{Additional Qualitative Results}
In this section, we provide additional qualitative results of the full bite acquisition pipeline: RetinaNet bounding box detection, SkeweringPoseNet for item pose estimation, ServoNet for pre-probing, and HapticVisualNet for primitive planning over consecutive actions. 

\begin{figure}[!htbp]
    \vspace{-0.3cm}
    \centering
    \includegraphics[width=1.0\textwidth]{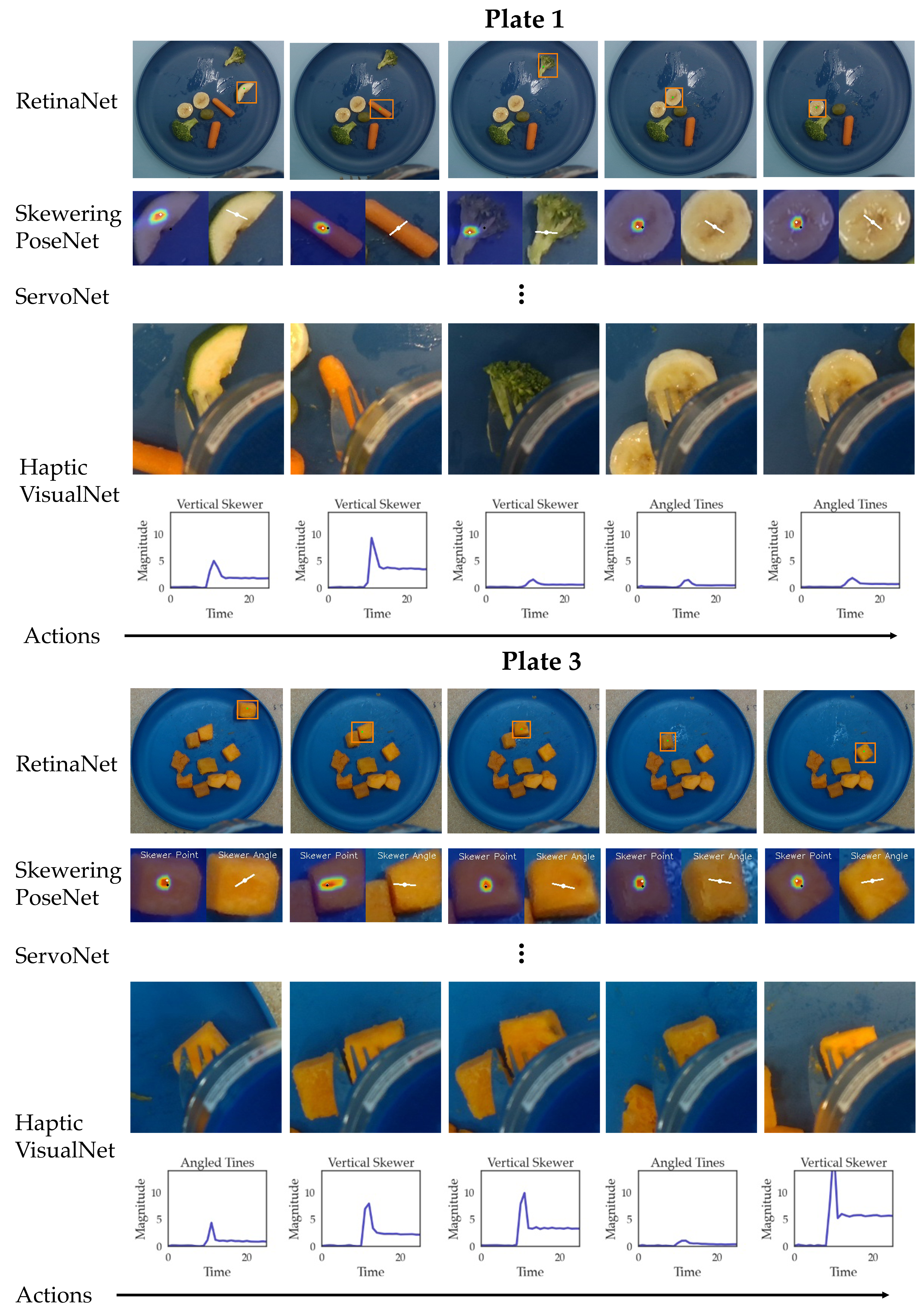}
    \caption{\textbf{HapticVis Action Predictions:} We visualize 5 consecutive HapticVis action predictions (left to right) for Plates 1 and Plates 3. The SkeweringPoseNet predictions per column first illustrate the predicted skewering points (which are intended to refine bounding box centers depicted in row 1 to item locations) and secondly the skewering orientations. In Plate 1, HapticVis correctly infers a sequence of actions to pick up zucchini, a raw carrot, a broccoli floret, and two slices of banana. In Plate 3, HapticVis skewers several pieces of butternut squash ranging from raw (1st-3rd, 5th columns) to overcooked (4th column). These intrinstic properties mostly align with the predicted action predictions, except for one mis-predicted \texttt{angled skewer} primitive for the first action, which succeeds but results in an unstable skewer. Plate 3 column 2 is a failed action in which ServoNet erroneously guides the fork to the wrong location, likely due to the close proximity of two butternut squash pieces. This causes a drop failure, but the item is re-attempted and successfully skewered in the next action.}
    \label{fig:action_preds_plate1}
\end{figure}

\section{ServoNet Training Details}
\label{sec:servonet_details}
We employ ServoNet, a network which continuously estimates the fork-food offset from images, and precisely guides the fork to a desired item via visual servoing in order to probe. We implement ServoNet with a fully-convolutional ResNet-18 backbone. ServoNet takes as input a $200 \times 200 \times 3$ image from the end-effector mounted RealSense camera, and outputs a $200 \times 200 \times 2$ heatmap. The two channels represent a 2-d Gaussian heatmap centered around the predicted fork pixel and nearest food pixel, respectively. We take the predicted fork and food pixels to be the argmaxes of these heatmaps.

\begin{figure}[!htbp]
    \centering
    \includegraphics[width=1.0\textwidth]{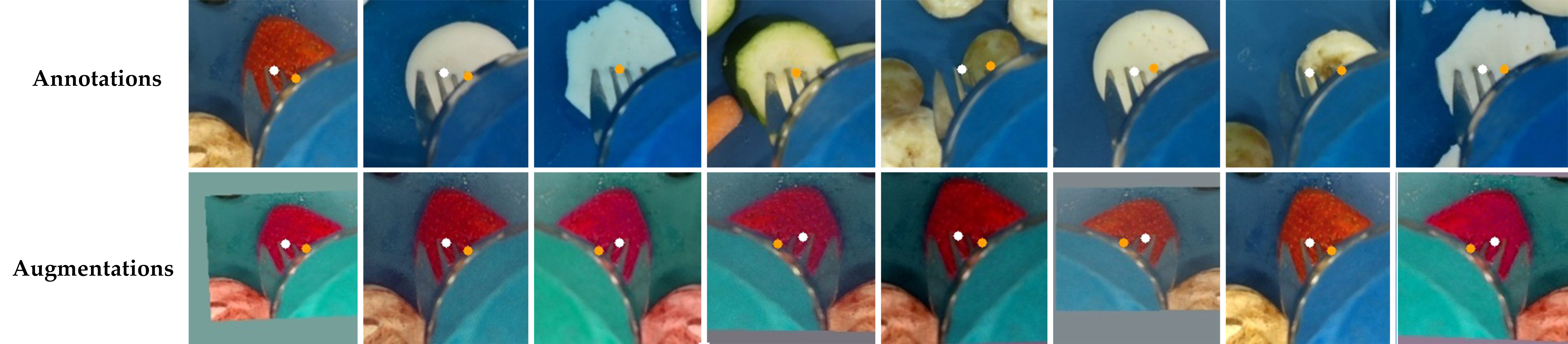}
    \caption{\textbf{ServoNet Dataset Generation:} Top row: we annotate 200 images with the fork tines center (white) and nearest food item center (orange). Bottom row: we augment this dataset with various colorspace and affine transformations to yield a dataset of 3,500 examples (shown for the first strawberry example).}
    \label{fig:servonet_annots}
\end{figure}

\begin{figure}[!htbp]
    \centering
    \includegraphics[width=1.0\textwidth]{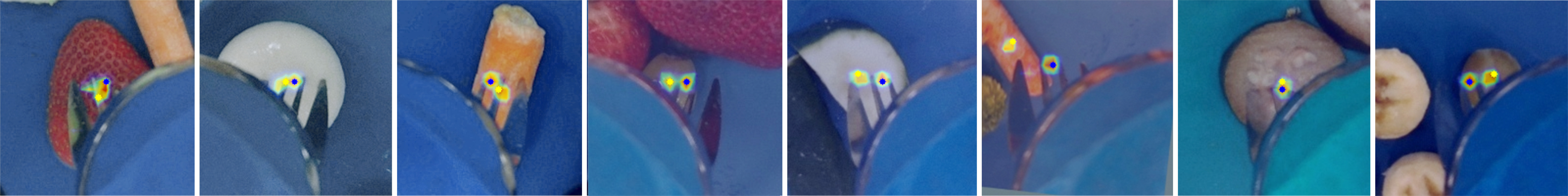}
    \caption{\textbf{ServoNet Predictions:} For 8 unseen images, we visualize the ServoNet Gaussian heatmap predictions. The estimated fork pixels appear in blue, and the estimated food item pixels are shown in yellow.}
    \label{fig:servonet_preds}
\end{figure}

To train ServoNet, we annotate 200 images with the corresponding fork and nearest food item pixels and augment this dataset to 3,500 examples (Figure \ref{fig:servonet_annots}). We train the network using binary cross-entropy loss over the predicted and ground truth heatmaps. Figure \ref{fig:servonet_preds} shows some visualizations of the predicted fork-food heatmaps on unseen images.

\section{Additional Ablations}
\label{sec:ablations_rebuttal}
\subsection{Sample Efficiency}
The original dataset collected to train HapticVisualNet consists of 300 examples of paired post-contact images, force readings, and manually assigned primitive labels, augmented 8x artificially (Section \ref{subsec:training_data}). To understand the sample efficiency of HapticVisualNet, we ablate for the accuracies of the network when training on varied amounts of data.

\begin{table}[!htbp]
\centering

 \begin{tabular}{c c c c} 

 Training Dataset Size (\% of orig.) & Overall Acc. & Vertical Skewer Acc. & Angled Skewer Acc. \\ 
 \hline
 $100\%$ (OURS) & $\mathbf{94.1\%}$ & $\mathbf{95.2\%}$ & $93.1\%$ \\
 $75\%$ & $91.8\%$ & $88.3\%$ & $95.7\%$ \\
 $50\%$ & $92.1\%$ & $87.6\%$ & $\mathbf{97.8\%}$ \\
 $25\%$ & $89.4\%$ & $83.4\%$ & $96.9\%$ \\
 $10\%$ & $84.2\%$ & $82.7\%$ & $86.1\%$ \\
\end{tabular}


\caption{ \textbf{HapticVisualNet Accuracy and Sample Efficiency:} Training with all 300 of the data points, augmented, yields the highest overall accuracy. We note a general trend towards lower accuracies as the dataset size decreases, taking into account that with smaller dataset sizes comes lower state coverage and higher variance in the accuracies, resulting in the 50\% network achieving high angled skewer accuracy but low vertical skewer accuracy.}
\label{tab:sample_eff_ablation}
\end{table}

\subsection{Interpretability of HapticVisualNet}

Given a paired food image observation and haptic readings, HapticVisualNet separately encodes visual features and haptic features. The concatenated features yield a 640-d multimodal embedding. This embedding serves as input to a final linear layer followed by a softmax activation which yields predicted primitive likelihoods. To improve interpretability of HapticVisualNet's learned food item representation, we visualize the multimodal embeddings for 26 items in the training dataset using 2-d t-SNE projections below. 

\begin{figure}[!htbp]
    \centering
    \includegraphics[width=0.8\textwidth]{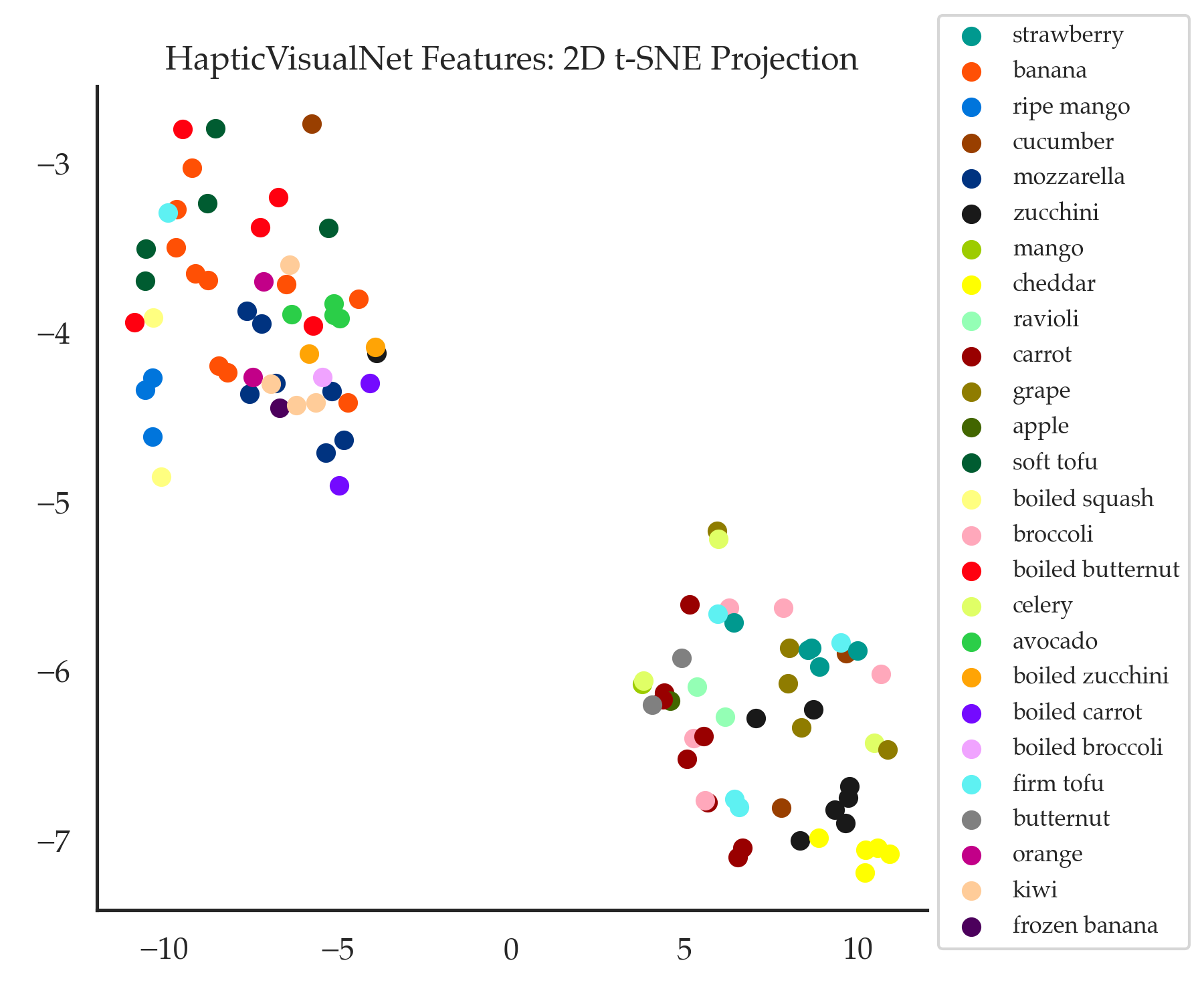}
    \caption{\textbf{2D t-SNE Projection of HapticVisualNet Embeddings}: We find that instances of the same item (i.e. ripe mango) and items with similar textural properties (i.e. boiled carrot and boiled zucchini, avocado and banana) tend to cluster. Additionally, HapticVisualNet learns to separate items that favor \texttt{vertical skewer} (bottom right) from those that favor \texttt{angled skewer} (upper left) by pushing them apart in latent space (i.e. mozzarella vs. cheddar, soft tofu vs. firm tofu).}
    \label{fig:tsne}
\end{figure}


\section{Additional Physical Experiments}
\label{sec:stress_tests}
In these set of experiments, we stress-test HapticVisualNet's capabilities to generalize to challenging textural and visual food properties. We assess HapticVisualNet on the tasks of  clearing a plate of frozen fruits and a plate with sauteed vegetables and tofu in sauce.

\begin{table}[!htbp]
\centering
 
 \begin{tabular}{c | c | c c c} 
 Items & Skewering Success Rate & Slip/Miss & BBox FP & 3+ Tries \\ 
 \hline
 Frozen mango, pineapple, strawberry & 20/25 \href{https://drive.google.com/file/d/1Cvv4XKfGlyfuzuAI_GgpIpFUbtcVX0Us/view}{[video]} & 5 & 0 & 1 \\
 Sautéed veggies and tofu & 20/23 \href{https://drive.google.com/file/d/1dWa7HlcuBp_V45Z7TJBb5xdMgpgKw6-A/view?usp=sharing}{[video]} & 3 & 3 & 0 \\ 
 Sautéed veggies, tofu, soy sauce & 25/34 \href{https://drive.google.com/file/d/1dWa7HlcuBp_V45Z7TJBb5xdMgpgKw6-A/view?usp=sharing}{[video]} & 9 & 6 & 1 \\ 
\end{tabular}


\caption{ \textbf{HapticVisualNet Stress-Tests:} HapticVisualNet successfully skewers frozen fruit in 20/25 and stir-fried tofu with vegetables in 45/57 total attempts. Videos of both experiments are available on the website. The failures observed in the fruit case are slips during probing due to the highly rigid texture of frozen fruits (and in the case of one strawberry, consecutively slipping three times on the same item). In this experiment, we also note that HapticVisualNet predominantly infers vertical skewering for these items as expected, but occasionally predicts angled tines. This is possibly due to the effects of thawing and softening over time. In the sautéed vegetables experiment, HapticVisualNet has more near misses and slips due to the degree of clutter and oiliness of the surface. As these items are very out of distribution, this also incurs more bounding box failures leading to early termination. Still, HapticVisualNet demonstrates generalization to charred/oiled/sauce-coated/seasoned items of varying levels of doneness. }
\label{tab:stress_tests}
\end{table}

\maketitle

%% file: 0-main.bbl
\begin{thebibliography}{32}
\providecommand{\natexlab}[1]{#1}
\providecommand{\url}[1]{\texttt{#1}}
\expandafter\ifx\csname urlstyle\endcsname\relax
  \providecommand{\doi}[1]{doi: #1}\else
  \providecommand{\doi}{doi: \begingroup \urlstyle{rm}\Url}\fi

\bibitem[Katz(1983)]{katz1983assessing}
S.~Katz.
\newblock Assessing self-maintenance: activities of daily living, mobility, and
  instrumental activities of daily living.
\newblock \emph{Journal of the American Geriatrics Society}, 1983.

\bibitem[Brault et~al.(2012)]{brault2012americans}
M.~W. Brault et~al.
\newblock \emph{Americans with disabilities: 2010}.
\newblock US Department of Commerce, Economics and Statistics Administration,
  US~…, 2012.

\bibitem[Brose et~al.(2010)Brose, Weber, Salatin, Grindle, Wang, Vazquez, and
  Cooper]{brose2010role}
S.~W. Brose, D.~J. Weber, B.~A. Salatin, G.~G. Grindle, H.~Wang, J.~J. Vazquez,
  and R.~A. Cooper.
\newblock The role of assistive robotics in the lives of persons with
  disability.
\newblock \emph{American Journal of Physical Medicine \& Rehabilitation},
  89\penalty0 (6):\penalty0 509--521, 2010.

\bibitem[Maheu et~al.(2011)Maheu, Archambault, Frappier, and
  Routhier]{maheu2011evaluation}
V.~Maheu, P.~S. Archambault, J.~Frappier, and F.~Routhier.
\newblock Evaluation of the jaco robotic arm: Clinico-economic study for
  powered wheelchair users with upper-extremity disabilities.
\newblock In \emph{2011 IEEE International Conference on Rehabilitation
  Robotics}, pages 1--5. IEEE, 2011.

\bibitem[obi()]{obi}
{Meet Obi}.
\newblock \url{https://meetobi.com/}.
\newblock [Online; accessed 6-June-2022].

\bibitem[mea()]{mealmate}
{Meal-Mate - Made2Aid}.
\newblock
  \url{https://www.made2aid.co.uk/productprofile?productId=8&company=RBF%20Healthcare&product=Meal-Mate}.
\newblock [Online; accessed 6-June-2022].

\bibitem[Gallenberger et~al.(2019)Gallenberger, Bhattacharjee, Kim, and
  Srinivasa]{gallenberger2019transfer}
D.~Gallenberger, T.~Bhattacharjee, Y.~Kim, and S.~S. Srinivasa.
\newblock Transfer depends on acquisition: Analyzing manipulation strategies
  for robotic feeding.
\newblock In \emph{2019 14th ACM/IEEE International Conference on Human-Robot
  Interaction (HRI)}, pages 267--276. IEEE, 2019.

\bibitem[Feng et~al.(2019)Feng, Kim, Lee, Gordon, Schmittle, Kumar,
  Bhattacharjee, and Srinivasa]{feng2019robot}
R.~Feng, Y.~Kim, G.~Lee, E.~K. Gordon, M.~Schmittle, S.~Kumar,
  T.~Bhattacharjee, and S.~S. Srinivasa.
\newblock Robot-assisted feeding: Generalizing skewering strategies across food
  items on a plate.
\newblock In \emph{The International Symposium of Robotics Research}, pages
  427--442. Springer, 2019.

\bibitem[Gordon et~al.(2020)Gordon, Meng, Bhattacharjee, Barnes, and
  Srinivasa]{gordon2020adaptive}
E.~K. Gordon, X.~Meng, T.~Bhattacharjee, M.~Barnes, and S.~S. Srinivasa.
\newblock Adaptive robot-assisted feeding: An online learning framework for
  acquiring previously unseen food items.
\newblock In \emph{2020 IEEE/RSJ International Conference on Intelligent Robots
  and Systems (IROS)}, pages 9659--9666. IEEE, 2020.

\bibitem[Gordon et~al.(2021)Gordon, Roychowdhury, Bhattacharjee, Jamieson, and
  Srinivasa]{gordon2021leveraging}
E.~K. Gordon, S.~Roychowdhury, T.~Bhattacharjee, K.~Jamieson, and S.~S.
  Srinivasa.
\newblock Leveraging post hoc context for faster learning in bandit settings
  with applications in robot-assisted feeding.
\newblock In \emph{2021 IEEE International Conference on Robotics and
  Automation (ICRA)}, pages 10528--10535. IEEE, 2021.

\bibitem[Bhattacharjee et~al.(2019{\natexlab{a}})Bhattacharjee, Lee, Song, and
  Srinivasa]{bhattacharjee2019role}
T.~Bhattacharjee, G.~Lee, H.~Song, and S.~S. Srinivasa.
\newblock Towards robotic feeding: Role of haptics in fork-based food
  manipulation.
\newblock \emph{IEEE Robotics and Automation Letters}, 4\penalty0 (2):\penalty0
  1485--1492, 2019{\natexlab{a}}.
\newblock \doi{10.1109/LRA.2019.2894592}.

\bibitem[Bhattacharjee et~al.(2019{\natexlab{b}})Bhattacharjee, Lee, Song, and
  Srinivasa]{bhattacharjee2019towards}
T.~Bhattacharjee, G.~Lee, H.~Song, and S.~S. Srinivasa.
\newblock Towards robotic feeding: Role of haptics in fork-based food
  manipulation.
\newblock \emph{IEEE Robotics and Automation Letters}, 4\penalty0 (2):\penalty0
  1485--1492, 2019{\natexlab{b}}.

\bibitem[Bhattacharjee et~al.(2020)Bhattacharjee, Gordon, Scalise, Cabrera,
  Caspi, Cakmak, and Srinivasa]{bhattacharjee2020more}
T.~Bhattacharjee, E.~K. Gordon, R.~Scalise, M.~E. Cabrera, A.~Caspi, M.~Cakmak,
  and S.~S. Srinivasa.
\newblock Is more autonomy always better? exploring preferences of users with
  mobility impairments in robot-assisted feeding.
\newblock In \emph{2020 15th ACM/IEEE International Conference on Human-Robot
  Interaction (HRI)}, pages 181--190. IEEE, 2020.

\bibitem[Belkhale et~al.(2022)Belkhale, Gordon, Chen, Srinivasa, Bhattacharjee,
  and Sadigh]{belkhale2022balancing}
S.~Belkhale, E.~K. Gordon, Y.~Chen, S.~Srinivasa, T.~Bhattacharjee, and
  D.~Sadigh.
\newblock Balancing efficiency and comfort in robot-assisted bite transfer.
\newblock In \emph{2022 International Conference on Robotics and Automation
  (ICRA)}, pages 4757--4763. IEEE, 2022.

\bibitem[Song et~al.(2019)Song, Bhattacharjee, and Srinivasa]{song2019sensing}
H.~Song, T.~Bhattacharjee, and S.~S. Srinivasa.
\newblock Sensing shear forces during food manipulation: resolving the
  trade-off between range and sensitivity.
\newblock In \emph{2019 International Conference on Robotics and Automation
  (ICRA)}, pages 8367--8373. IEEE, 2019.

\bibitem[Srivastava et~al.(2022)Srivastava, Li, Lingelbach,
  Mart{\'\i}n-Mart{\'\i}n, Xia, Vainio, Lian, Gokmen, Buch, Liu,
  et~al.]{srivastava2022behavior}
S.~Srivastava, C.~Li, M.~Lingelbach, R.~Mart{\'\i}n-Mart{\'\i}n, F.~Xia, K.~E.
  Vainio, Z.~Lian, C.~Gokmen, S.~Buch, K.~Liu, et~al.
\newblock Behavior: Benchmark for everyday household activities in virtual,
  interactive, and ecological environments.
\newblock In \emph{Conference on Robot Learning}, pages 477--490. PMLR, 2022.

\bibitem[Lin et~al.(2020)Lin, Wang, Olkin, and Held]{lin2020softgym}
X.~Lin, Y.~Wang, J.~Olkin, and D.~Held.
\newblock Softgym: Benchmarking deep reinforcement learning for deformable
  object manipulation.
\newblock 2020.

\bibitem[Suh and Tedrake(2020)]{suh2020surprising}
H.~Suh and R.~Tedrake.
\newblock The surprising effectiveness of linear models for visual foresight in
  object pile manipulation.
\newblock In \emph{International Workshop on the Algorithmic Foundations of
  Robotics}, pages 347--363. Springer, 2020.

\bibitem[Erickson et~al.(2020)Erickson, Gangaram, Kapusta, Liu, and
  Kemp]{erickson2020assistive}
Z.~Erickson, V.~Gangaram, A.~Kapusta, C.~K. Liu, and C.~C. Kemp.
\newblock Assistive gym: A physics simulation framework for assistive robotics.
\newblock In \emph{2020 IEEE International Conference on Robotics and
  Automation (ICRA)}, pages 10169--10176. IEEE, 2020.

\bibitem[Gemici and Saxena(2014)]{gemici2014learning}
M.~C. Gemici and A.~Saxena.
\newblock Learning haptic representation for manipulating deformable food
  objects.
\newblock In \emph{2014 IEEE/RSJ International Conference on Intelligent Robots
  and Systems}, pages 638--645. IEEE, 2014.

\bibitem[Zhang et~al.(2019)Zhang, Sharma, Veloso, and
  Kroemer]{zhang2019leveraging}
K.~Zhang, M.~Sharma, M.~Veloso, and O.~Kroemer.
\newblock Leveraging multimodal haptic sensory data for robust cutting.
\newblock In \emph{2019 IEEE-RAS 19th International Conference on Humanoid
  Robots (Humanoids)}, pages 409--416. IEEE, 2019.

\bibitem[Sawhney et~al.(2020)Sawhney, Lee, Zhang, Veloso, and
  Kroemer]{sawhney2020playing}
A.~Sawhney, S.~Lee, K.~Zhang, M.~Veloso, and O.~Kroemer.
\newblock Playing with food: Learning food item representations through
  interactive exploration.
\newblock In \emph{International Symposium on Experimental Robotics}, pages
  309--322. Springer, 2020.

\bibitem[Sharma et~al.(2019)Sharma, Zhang, and Kroemer]{sharma2019learning}
M.~Sharma, K.~Zhang, and O.~Kroemer.
\newblock Learning semantic embedding spaces for slicing vegetables.
\newblock \emph{arXiv preprint arXiv:1904.00303}, 2019.

\bibitem[Heiden et~al.(2022)Heiden, Macklin, Narang, Fox, Garg, and
  Ramos]{heiden2022disect}
E.~Heiden, M.~Macklin, Y.~Narang, D.~Fox, A.~Garg, and F.~Ramos.
\newblock Disect: A differentiable simulator for parameter inference and
  control in robotic cutting.
\newblock \emph{arXiv preprint arXiv:2203.10263}, 2022.

\bibitem[Zhang et~al.(2019)Zhang, Sharma, Veloso, and
  Kroemer]{zhang2019cutting}
K.~Zhang, M.~Sharma, M.~Veloso, and O.~Kroemer.
\newblock Leveraging multimodal haptic sensory data for robust cutting.
\newblock \emph{CoRR}, abs/1909.12460, 2019.
\newblock URL \url{http://arxiv.org/abs/1909.12460}.

\bibitem[Mahler et~al.(2017)Mahler, Liang, Niyaz, Laskey, Doan, Liu, Ojea, and
  Goldberg]{mahler2017dex}
J.~Mahler, J.~Liang, S.~Niyaz, M.~Laskey, R.~Doan, X.~Liu, J.~A. Ojea, and
  K.~Goldberg.
\newblock Dex-net 2.0: Deep learning to plan robust grasps with synthetic point
  clouds and analytic grasp metrics.
\newblock 2017.

\bibitem[Florence et~al.(2018)Florence, Manuelli, and
  Tedrake]{florence2018dense}
P.~R. Florence, L.~Manuelli, and R.~Tedrake.
\newblock Dense object nets: Learning dense visual object descriptors by and
  for robotic manipulation.
\newblock 2018.

\bibitem[Ha and Song(2022)]{ha2022flingbot}
H.~Ha and S.~Song.
\newblock Flingbot: The unreasonable effectiveness of dynamic manipulation for
  cloth unfolding.
\newblock In \emph{Conference on Robot Learning}, pages 24--33. PMLR, 2022.

\bibitem[Lee et~al.(2019)Lee, Zhu, Srinivasan, Shah, Savarese, Fei-Fei, Garg,
  and Bohg]{lee2019making}
M.~A. Lee, Y.~Zhu, K.~Srinivasan, P.~Shah, S.~Savarese, L.~Fei-Fei, A.~Garg,
  and J.~Bohg.
\newblock Making sense of vision and touch: Self-supervised learning of
  multimodal representations for contact-rich tasks.
\newblock In \emph{2019 International Conference on Robotics and Automation
  (ICRA)}, pages 8943--8950. IEEE, 2019.

\bibitem[Fazeli et~al.(2019)Fazeli, Oller, Wu, Wu, Tenenbaum, and
  Rodriguez]{fazeli2019see}
N.~Fazeli, M.~Oller, J.~Wu, Z.~Wu, J.~B. Tenenbaum, and A.~Rodriguez.
\newblock See, feel, act: Hierarchical learning for complex manipulation skills
  with multisensory fusion.
\newblock \emph{Science Robotics}, 4\penalty0 (26):\penalty0 eaav3123, 2019.

\bibitem[Li et~al.(2019)Li, Zhu, Tedrake, and Torralba]{li2019connecting}
Y.~Li, J.-Y. Zhu, R.~Tedrake, and A.~Torralba.
\newblock Connecting touch and vision via cross-modal prediction.
\newblock In \emph{Proceedings of the IEEE/CVF Conference on Computer Vision
  and Pattern Recognition}, pages 10609--10618, 2019.

\bibitem[Lee et~al.(2021)Lee, Tan, Zhu, and Bohg]{lee2021detect}
M.~A. Lee, M.~Tan, Y.~Zhu, and J.~Bohg.
\newblock Detect, reject, correct: Crossmodal compensation of corrupted
  sensors.
\newblock In \emph{2021 IEEE International Conference on Robotics and
  Automation (ICRA)}, pages 909--916. IEEE, 2021.

\end{thebibliography}
